\documentclass[letterpaper,10pt,conference]{ieeeconf}

\IEEEoverridecommandlockouts                              

\usepackage{times}
\usepackage{epsfig}
\usepackage{graphicx}
\usepackage{amsmath}
\usepackage{amssymb}
\usepackage{xspace}

\usepackage{amsfonts}

\usepackage{epstopdf}

\usepackage{multirow}
\usepackage{pbox}

\usepackage[pagebackref=true,breaklinks=true,letterpaper=true,colorlinks,bookmarks=false]{hyperref}

\title{\LARGE \bf
Combined Image- and World-Space Tracking in Traffic Scenes
}

\author{Aljo\v{s}a O\v{s}ep, Wolfgang Mehner, Markus Mathias, and Bastian Leibe
\thanks{\scriptsize The authors are with the Visual Computing Institute, RWTH Aachen University
\newline
        E-mail: {\tt\scriptsize \{osep, mehner, mathias, leibe\}@vision.rwth-aachen.de}}%
}

\newcommand{\PAR}[1]{\vskip4pt \noindent {\bf #1~}}

\makeatletter
\DeclareRobustCommand\onedot{\futurelet\@let@token\@onedot}
\def\@onedot{\ifx\@let@token.\else.\null\fi\xspace}

\def\eg{\emph{e.g}\onedot} 
\def\ie{\emph{i.e}\onedot}

\def\wrt{w.r.t\onedot} 
\def\etal{\emph{et al}\onedot}

\def\Sec{Sec\onedot}
\def\Fig{Fig\onedot}
\makeatother

\DeclareMathOperator*{\argmin}{\arg\!\min}

\DeclareMathOperator{\hypopose}{\mathbf{p}}
\DeclareMathOperator{\hypovelocity}{\mathbf{v}}
\DeclareMathOperator{\hyposize}{\mathbf{s}}
\DeclareMathOperator{\hypobbox_2d}{\mathbf{b}^{2D}}
\DeclareMathOperator{\hypobvelo_2d}{\mathbf{\dot{b}}^{2D}}
\DeclareMathOperator{\hypocategory}{\mathbf{c}}

\DeclareMathOperator{\hypoinliers}{\mathcal{I}}

\newcommand{\binvec}{\ensuremath{\mathbf{m}}}

\newcommand{\detection}[2]{\ensuremath{m_{det}^{#1, #2}}}

\newcommand{\proposal}[2]{\ensuremath{m_{prop}^{#1, #2}}}

\newcommand{\observset}[2]{\ensuremath{\operatorname{\mathbf{o}^{#1}_{#2}}}}
\newcommand{\obs}[2]{\ensuremath{\operatorname{o^{#1}_{#2}}}}

\newcommand{\hyposet}{\ensuremath{\operatorname{\mathbf{h}}}}
\newcommand{\hypo}[2]{\ensuremath{\operatorname{h^{#1}_{#2}}}}

\begin{document}

\maketitle
\thispagestyle{empty}
\pagestyle{empty}

\begin{abstract}
Tracking in urban street scenes plays a central role in autonomous systems such as self-driving cars.
Most of the current vision-based tracking methods perform tracking in the image domain.
Other approaches, \eg based on LIDAR and radar, track purely in 3D.
While some vision-based tracking methods invoke 3D information in parts of their pipeline,
and some 3D-based methods utilize image-based information in components of their approach,
we propose to use image- and world-space information jointly throughout our method.
We present our tracking pipeline as a 3D extension of image-based tracking.
From enhancing the detections with 3D measurements to the reported positions of every tracked object,
we use world-space 3D information at every stage of processing.
We accomplish this by our novel coupled 2D-3D Kalman filter,
combined with a conceptually clean and extendable hypothesize-and-select framework.
Our approach matches the current state-of-the-art on the official KITTI benchmark, which performs evaluation in the 2D image domain only.
Further experiments show significant improvements in 3D localization precision by enabling our coupled 2D-3D tracking.
\end{abstract}
\section{Introduction}
\label{sec:intro}
Visual scene understanding in outdoor environments is a key requirement for autonomous mobile systems.
The tracking and detection of traffic participants such as pedestrians, cars, and bicyclists
plays an important role in safe navigation of autonomous vehicles.
Through tracking, the vehicle becomes aware of the whereabouts of important objects and determines their motion.

A substantial amount of research has been done in this area,
mainly driven by the goal of developing autonomous vehicles that may operate in everyday traffic.
Recent advances in object detection \cite{Wang13ICCV, Chen15NIPS, Xiang15CVPR}
and detection-based multi-object tracking \cite{Choi15ICCV, Xiang15ICCV, Lenz15ICCV, Yoon16CVPR}
start to approach a matured state.
However, there are still several open problems in vision-based tracking approaches.
The majority of existing methods only perform tracking in the image domain.
Yet, in mobile robotics and autonomous driving scenarios,
precise 3D localization and trajectory estimation is of fundamental importance. In order to prevent collisions, 
it is crucial to be aware of the extent and the orientation of objects in world-space, especially for objects close to the camera.

\begin{figure}[t]
\includegraphics[width=1.0\linewidth]{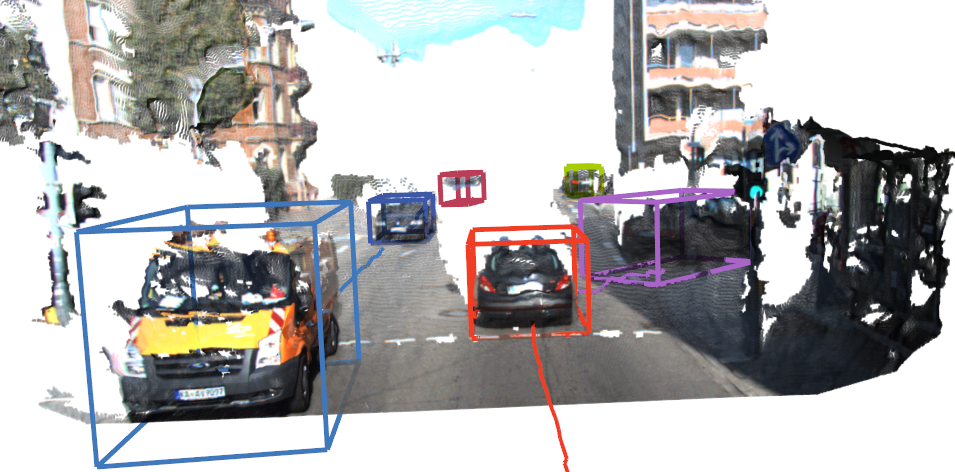}
\vspace*{-8mm}
\caption{Example output of our method. Using stereo matching and vision-based tracking, we obtain precise 3D bounding boxes.}
\label{fig:teaser}
\end{figure}

In this work, we carefully combine 2D object detections and 3D stereo depth measurements in order to improve image-based tracking
and, more importantly, precise 3D localization (see Fig.~\ref{fig:teaser}).
While image-based tracking has shown to be successful even in greater distances from the camera, 3D stereo measurement precision deteriorates quickly with camera distance \cite{Pinggera14ECCV}.
Our system weights these sources depending on the distance from the camera.
It combines 2D and 3D information when available,
but is also able to cope with missing 3D measurements.
Our contributions are as follows. 
(i) We propose a new tracking framework\footnote{Code is available at \scriptsize http://www.vision.rwth-aachen.de/page/combined-tracking}, which exploits both 2D and 3D measurements.
To that end, we combine object detections (\eg cars, pedestrians) and 3D object proposals obtained in a 3D point-cloud.
Our method takes advantage of the strengths of both sources of information: 2D detections provide class information, while our 3D proposals assist in locating the objects in world coordinates.
(ii) We introduce a novel 2D-3D Kalman filter, which is keeping both an image- and a world-space (position and size) estimate.
These estimates are loosely coupled to ensure the consistency of a track.
This coupling enables us to track distant objects and continue these tracks with more precise information in the close range,
while smoothly transitioning between the modalities.
(iii) We show competitive results on the KITTI benchmark. Additionally to these image-based evaluations, we assess the precision of our method in 3D space to quantify the advantage of our method.

\section{Related Work}
\label{sec:relatedWork}

\PAR{Vision-based multi-object tracking in street scenes.}
Most vision-based approaches to multi-object tracking (MOT) in street scenes follow the tracking-by-detection paradigm, where object detector responses are matched across multiple frames in the image domain \cite{Choi15ICCV,Choi13PAMI,Lenz15ICCV, Milan14TPAMI, Pirsiavash11CVPR, Xiang15ICCV, Yoon16CVPR,Zhang08CVPR}.
Geiger \etal \cite{Geiger2014PAMI} associate detections using an appearance model and bounding box overlap in the image domain.
Detection bounding boxes are filtered with a Kalman filter and are associated in a two stage process: first, detections are combined to short tracks (tracklets), followed by association of these tracklets to form full trajectories.
Yoon \etal \cite{Yoon16CVPR} propose a method that compensates for abrupt camera motion by employing a new data association algorithm that takes structural constraints into account.
Choi \cite{Choi15ICCV} utilizes a sparse optical flow-based descriptor as an additional affinity measure and proposes a near-online tracking formulation similar to \cite{Leibe08TPAMI}.
These methods are based on monocular camera systems and perform tracking in the image domain.
However, for robotics applications such as path planning and obstacle avoidance it is desirable to be fully aware of the object's 3D position and spatial extent.
Our method follows the same tracking-by-detection paradigm, but performs tracking jointly in 2D and 3D.

\PAR{Vision-based multi-object tracking using depth sensors.} Several vision-based tracking methods include depth information in the tracking process by using either stereo camera pairs or structured light based (RGB-D) sensors (\eg Kinect, RealSense) in order to improve tracking performance. Depth information can be exploited for different purposes, \eg to enhance the detection process \cite{Beymer99, Jafari14ICRA, Zhang13}, to reduce a detector's search space \cite{Bansal10ICRA, Beymer99, Jafari14ICRA, Mitzel11ICCVW, Mitzel11BMVC}, or to facilitate data association \cite{Ess09PAMI,Leibe08TPAMI}.
While \mbox{RGB-D} sensors are well suited to perform pedestrian tracking in indoor scenes \cite{Beymer99, Jafari14ICRA, Zhang13} they are of less use in outdoor environments due to challenging lighting conditions and reflective surfaces.
Therefore, multi-object tracking systems in street scenes usually rely on a stereo-camera based setup.
\cite{Ess09PAMI, Leibe08TPAMI, Mitzel10ECCV} use such a setup to estimate a coarse scene geometry (ground plane) and localize detections in 3D-space by ray-casting detection footpoints and intersecting them with the ground plane \cite{Leibe08TPAMI} or by performing a depth-analysis of the detection windows \cite{Ess09PAMI, Mitzel10ECCV}.
While such a depth-analysis of detection windows has shown to work reliably for the pedestrian category, it can fail in more challenging scenarios, \eg for occluded cars.

Similar to our work, \cite{Song15CVPR} precisely estimates 3D pose and extent of the tracked objects. However, their method is limited to cars. We show the applicability of our approach for a multitude of object categories.
For a detailed overview of MOT methods using RGB-D sensors we refer the interested reader to \cite{Camplani16arxiv}.

	
\PAR{LIDAR-based multi-object tracking.}
When performing MOT using LIDAR sensors, the tracking pipeline is typically reversed using a tracking-before-detection paradigm \cite{Dewan15ICRA, Kaestner12ICRA, Moosmann13ICRA, Teichman11ICRA}.
LIDAR sensor outputs are more precise, and do not suffer from systematic errors compared to stereo-based sensors.
The acquired measurements are better suitable to delineate the shapes of objects \cite{Cho14ICRA}.
Segmentation of object candidates from LIDAR sensor data is a well-studied problem \cite{Moosmann13ICRA, Teichman11ICRA, Wang12ICRA}.
These object candidates provide a precise object boundary, position and shape but no object category information, hence model-free tracking is performed (typically, using a Kalman filter and nearest-neighbor data association).
Category-agnostic trajectories can then be classified into object categories \cite{Teichman11ICRA, Wang12ICRA}.
In our work, we also rely on category-agnostic object proposals, but rather than using expensive LIDAR systems our proposals are generated from stereo input images~\cite{Osep16ICRA}.

In the context of LIDAR tracking, it has been shown that 3D measurements of the objects can be utilized to improve 3D tracking precision (position and velocity).
\cite{Held14RSS} proposes to align LIDAR object candidates in a coarse-to-fine fashion using annealed dynamic histograms in order to obtain precise position and velocity.
\cite{Ushani15IROS} uses 3D measurements to jointly estimate trajectory and shape of the tracked object by creating an object ``map''.
We show that using 3D measurements is suitable for improving the 3D tracking precision even though our input data originates from noisy stereo-based depth data.

%

\section{Method Overview}
\label{sec:overview}

\begin{figure}[t]
\includegraphics[width=1.0\linewidth]{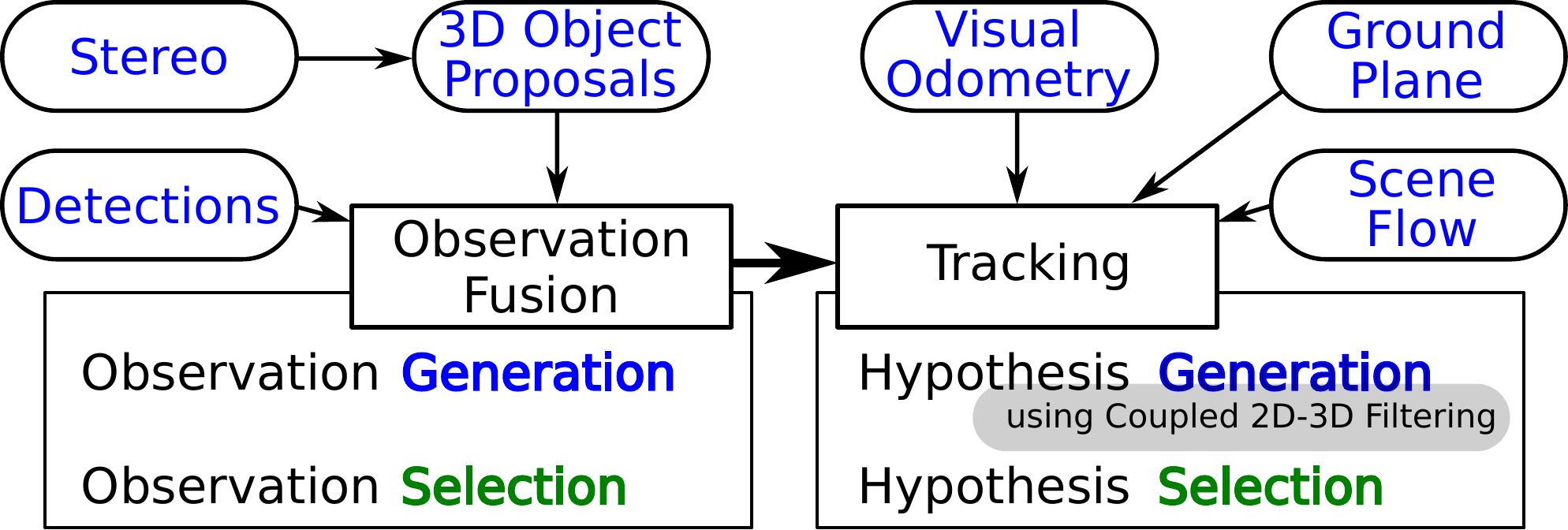}
\caption{Method overview.
	Our input data is processed in two steps.
	We first generate a large number of observations (observation = detection + 3D~object~proposal) and perform model selection to pick the most suitable ones.
	In the same fashion, we track the observations to generate an over-complete set of tracking hypotheses,
	and again perform model selection to pick the trajectories we report as results.
	}
\label{fig:pipeline}
\end{figure}

\begin{figure*}[t!]

\newlength{\figBBimage}
\newlength{\figBBcloud}
\setlength{\figBBimage}{0.4\linewidth}
\setlength{\figBBcloud}{0.4\linewidth}

\begin{center}
\includegraphics[width=\figBBimage]{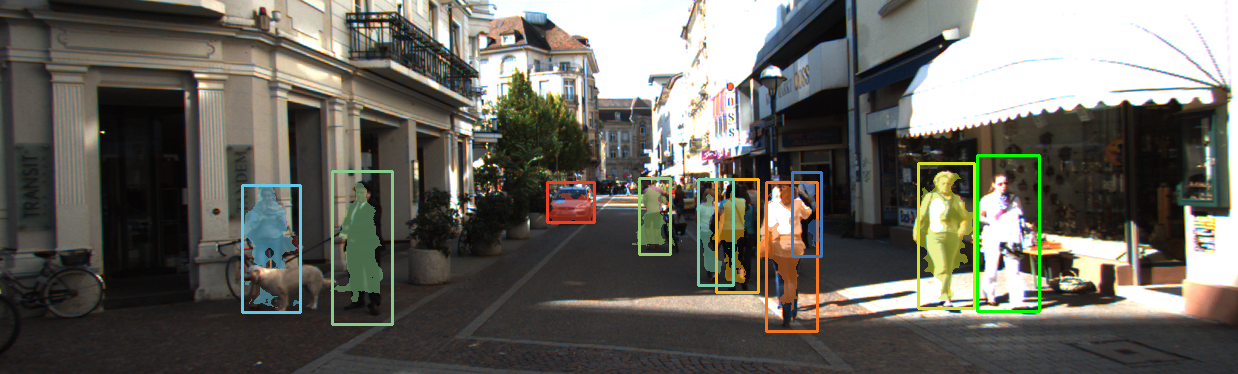}
\includegraphics[width=\figBBimage]{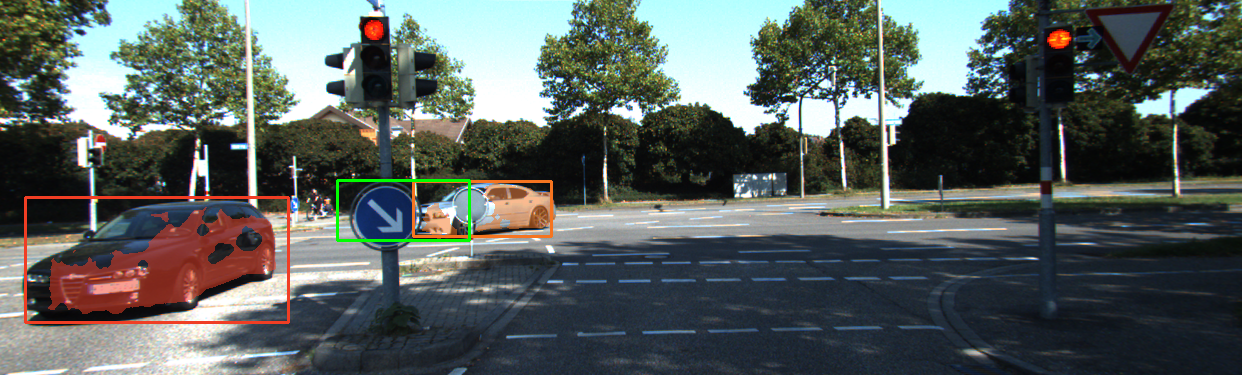}

\includegraphics[width=\figBBcloud]{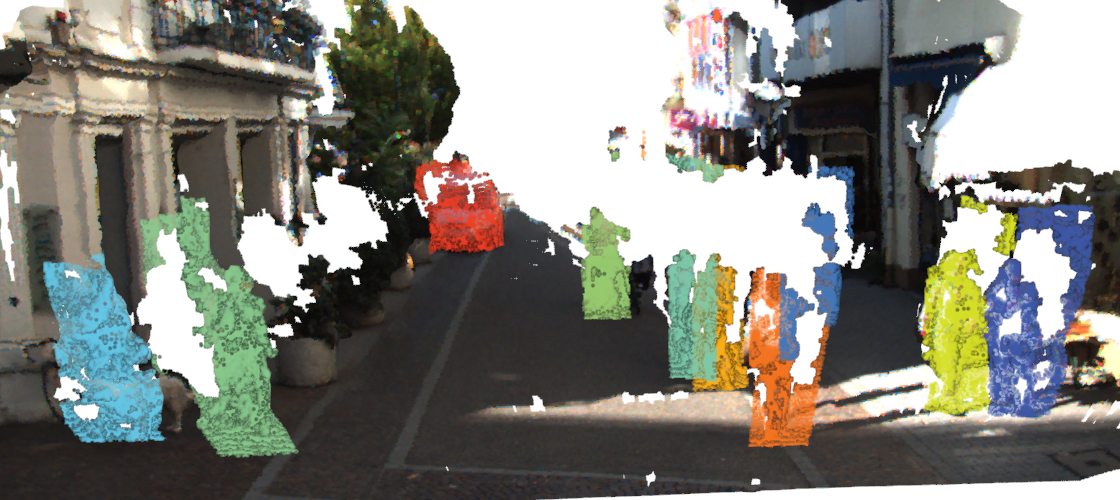}
\includegraphics[width=\figBBcloud]{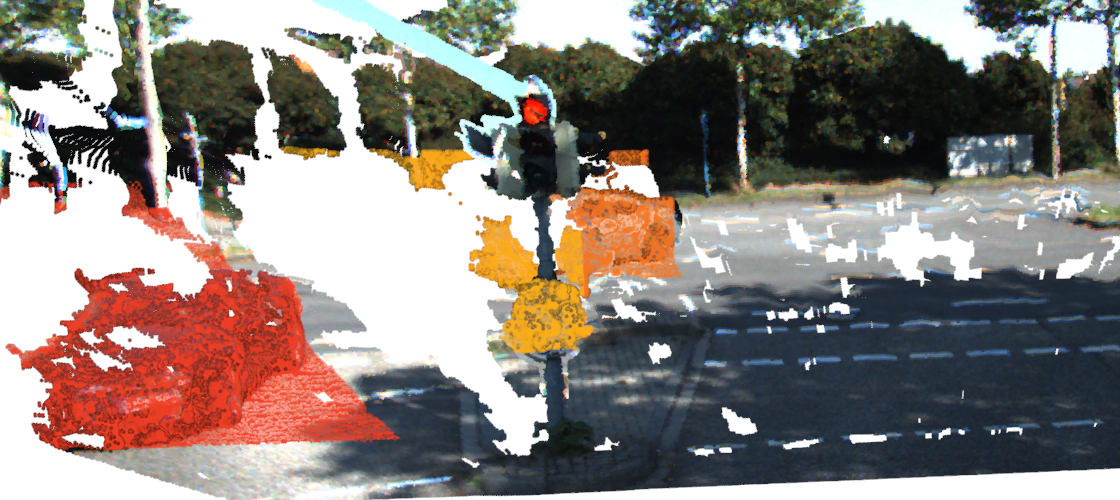}

\includegraphics[width=\figBBcloud]{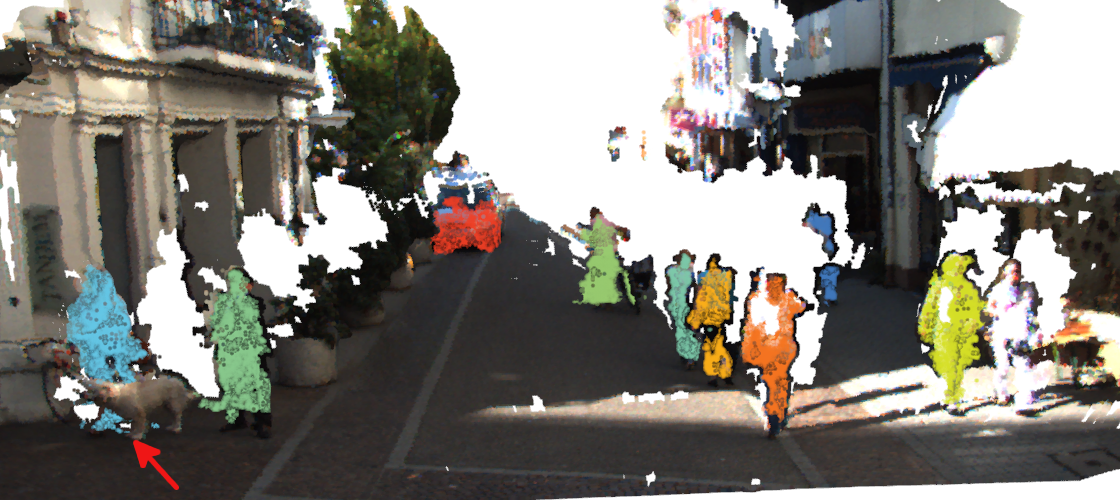}
\includegraphics[width=\figBBcloud]{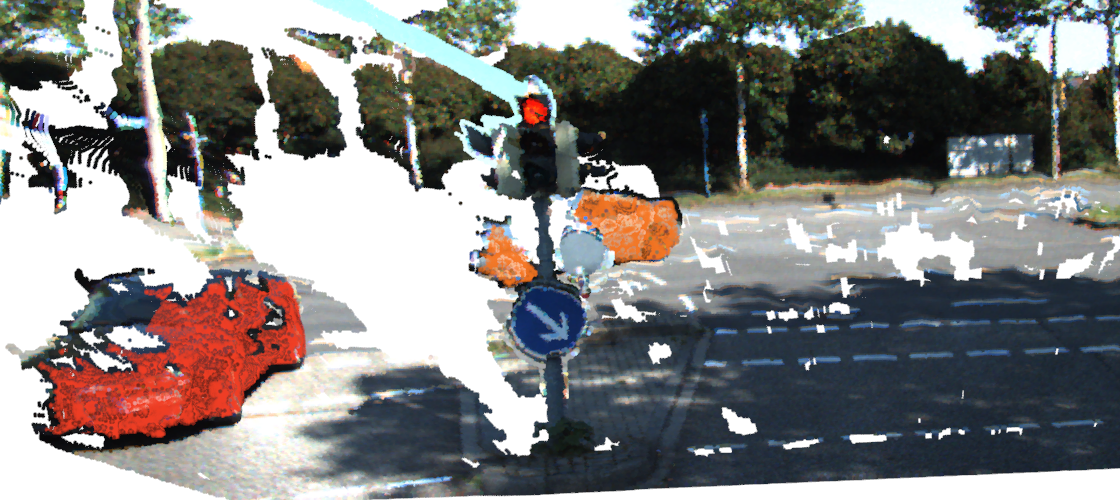}
\end{center}
\caption{Use of depth information. The top row shows detections and associated 3D object proposals (bright green boxes are not associated).
	The middle row shows the corresponding stereo point-clouds, where all points originating from inside the bounding boxes are highlighted.
	These sets of points often include an occluder or the background.
	The bottom row shows the same point-clouds, but points belonging to the associated 3D proposals are highlighted.
	The 3D proposals separate the object from the environment much better, \eg the dog in the left column.
	\textit{Best viewed in color.}}
\label{fig:bb_and_3d_proposals}
\end{figure*}

\Fig~\ref{fig:pipeline} shows an overview of our proposed pipeline.
Our method combines information from several commonly used sources,
such as object detections \cite{Wang13ICCV, Geiger2014PAMI}, stereo \cite{Geiger10ACCV}, visual odometry \cite{Geiger11IV}, and optionally scene flow \cite{Vogel13ICCV}.
Additionally, we support the detections with stereo-based class-agnostic 3D object proposals \cite{Osep16ICRA}, short \textit{3D proposals}.
These proposals hypothesize objects and provide precise 3D measurements for them, in our case the objects' positions on the ground plane, physical size and a segmentation mask (See Fig. \ref{fig:bb_and_3d_proposals}, bottom).

A 3D object proposal combined with a 2D detection constitutes an \textit{observation}.
We use a CRF model to select suitable observations out of the huge set of possible observations we generate.
The CRF scores the observations according to the compatibility of their 2D and 3D information,
and helps to exclude observations which would share either a detection or a 3D proposal with another selected observation.

Our tracker works using a comparable idea.
We first generate an over-complete set of track hypotheses.
They are then selected via a CRF model,
which scores the hypotheses and prevents the selection of overlapping pairs of hypotheses.

In extension of the common paradigm of image-space tracking, we use 3D information in every part of the tracking pipeline.
Hypotheses are tracked using Kalman filters with a joint 2D-3D state, which is weakly coupled by projection and back-projection operations.
As a result, measurements of 2D bounding boxes can help to estimate the 3D position and vice versa.
This also means that we can track \textit{opportunistically}, \ie we make use of the 3D measurements when available,
but we can also perform the tracking without them.
Finally, the CRF model scores the hypotheses by evaluating their consistency in image- and world-space.

\section{Observation Fusion Model}

As inputs to our observation fusion we use 3D object proposals, obtained by performing clustering in the stereo point-cloud, and object detections.
We fuse these sources of information before feeding them to the tracking process.
This results in (i) extending the 2D detections by precise 3D measurements of the object position and size
and (ii) selecting the relevant proposals from the huge set of available 3D proposals.

\subsection{Observation Models}
\label{sec:observation_models}

For 2D object detection we use image-space detectors, \eg \cite{Wang13ICCV, Geiger2014PAMI}.
A set of detections at timestep $t$, $t \in \left [ 0, \dots, T\right ]$, is defined as $\mathcal{M}_{det}^{t} = \left \{ m_{det}^{t,i} \right \}, i \in 1, \ldots, n_t$.
Each detection measurement provides
the center-point, width, and height of the 2D bounding box $ \hypobbox_2d = \left[ x^{2D}, y^{2D}, w^{2D}, h^{2D} \right]^T $ (in the image domain),
class information $c$, and score $s_{\text{det}}$:
\begin{eqnarray} 
\label{equ:detection_obs}
\detection{t}{i} = \left [ \hypobbox_2d, c, s_{\text{det}} \right ]^T .
\end{eqnarray}

We obtain 3D object proposals using an extension of our previous work on multi-scale 3D proposals, described in \cite{Osep16ICRA}.
That approach generates a large set of class-agnostic 3D object proposals by identifying clusters of depth measurements in stereo point-clouds.
The intuition behind this approach is that relevant real-world objects usually stick out of the ground plane, surrounded by a certain amount of free space.
As the class of each object is unknown, potential objects have to be searched at varying scales. Clusters that remain stable over multiple scales are selected as possible object candidates.
The number of proposed objects is reduced by merging clusters if their bounding boxes overlap ($\text{intersection-over-union} > 0.9$).

For our application, we slightly modify \cite{Osep16ICRA}:
(i) Rather than using the projected bounding boxes in the image plane for merging the clusters, we now project 3D bounding boxes to the ground plane. This results in more precise localization in 3D.
(ii) Additionally to clustering with an isotropic kernel, we add two anisotropic kernels elongated along the x- and z-direction (the two dimensions of the ground plane) in order to better represent elongated objects.
As a result, we obtain a rich set of 3D object proposals $\mathcal{M}_{prop}^{t} = \left \{ m_{prop}^{t,j} \right \}, j \in 1, \ldots, m_t$.
Each proposal is defined by its position $ \hypopose = \left[ x, y, z \right]^T $, velocity $ \hypovelocity = \left[ \dot{x}, \dot{y}, \dot{z} \right]^T $ (obtained from scene flow \cite{Vogel13ICCV}),
size estimate $ \hyposize = \left[ w^{3D}, h^{3D}, l^{3D} \right]^T $, and score:
\begin{eqnarray} 
\label{equ:proposal_obs}
\proposal{t}{j} = \left [ \hypopose, \hypovelocity, \hyposize, s_{\text{prop}} \right ]^T.
\end{eqnarray}


\subsection{Observation Fusion}
\label{subsec:observation_fusion}

At this step, we have a set of detections, which provide object class information,
and a set of 3D proposals, which provide localization and object size estimates.
However, no category information is associated to these proposals.
Therefore, a group of pedestrians is a valid object, as much as each pedestrian individually is.

Only by picking a proposal at the right scale, its measurement becomes meaningful semantically.
This is a first instance of image- and world-space information supporting each other.
A detection has no precise 3D information,
which is obtained by associating the detection to a proposal.
On the other hand, a 3D measurement can only be considered precise (in a semantic sense) if the proposal is derived from a segment of the point-cloud which corresponds to an actual object.
The set of detections helps selecting meaningful proposals, and rejecting the others.
See \Fig~\ref{fig:bb_and_3d_proposals} for examples of associated detections and proposals.

We address the selection problem by performing MAP inference in a CRF model.
We enumerate a large set of possible associations between detections and 3D proposals.
In theory, we would need to enumerate the Cartesian product of both sets to obtain the overlapping observation set $\observset{t}{} = \mathcal{M}_{det}^{t} \times \mathcal{M}_{prop}^{t}$.
In practice we do gating and have only a limited number of overlapping observations.
These associations are interpreted as nodes in a graph and we assign binary labels $\mathbf{s} \in \{0,1\}^{|\observset{}{}|}$ (1: selected, 0: not selected).
By minimizing the following energy, we obtain a set of consistent associations between the proposals and the detections.
%
\begin{equation}
\label{equ:energy_observation_model}
\mathbf{E}(\mathbf{s}, \observset{}{}) = \sum_{o_i \in \observset{}{}} s_i \phi \left( o_i \right) + \sum_{o_i,o_j \in \observset{}{}} s_i s_j \varphi \left( {o_i, o_j} \right ),
\end{equation}
where the detection-to-proposal association potential is defined as:
\begin{eqnarray}
\label{equ:energy_observation_model_unary}
\phi \left( \obs{}{i} \right) = - w^o_1 \phi_{\text{size}} \left ( \obs{}{i} \right ) -  w^o_2 \phi_{\text{pos}} \left ( \obs{}{i} \right ) - w^o_3 \phi_{\text{proj}} \left ( \obs{}{i} \right ) + w^o_4
\end{eqnarray}
The first term $ \phi_{\text{size}} \left ( \obs{}{i} \right ) $ scores the 3D proposal size, using the probability of the size given statistics (mean+variance) learned from data.
The second term ensures that the detection and the proposal have a small distance on the ground plane (Mahalanobis distance given the uncertainty of the measurements).
The third term matches the projected area of the proposal with the area of the detection bounding box (intersection-over-union).
The weight $w^o_4$ imposes a minimal requirement on the association,
since observations with an overall positive score $ \phi \left( \obs{}{i} \right) $ will not be selected.

The pairwise term penalizes overlapping associations:
\begin{equation}
\varphi \left( {o_i, o_j} \right ) = w^o_5 \cdot \frac{\left | P_i \bigcap  P_j \right | }{\text{min} \left( \left | P_i \right | , \left | P_j \right | \right )}
+ w^o_6 \cdot \mathbb{I} \left ( o_i, o_j\right ) .
\end{equation}
The first term measures the (normalized) overlap of the two observations $i$ and $j$,
based on the number of shared 3D points $| P_i\bigcap P_j |$ of their 3D proposals.
The second term is a hard-exclusion term: the indicator function $\mathbb{I} \left ( o_i, o_j \right )$ is $1$ if two observation share a proposal or a detection and $0$ otherwise.
The purpose of the pairwise term is to penalize physical overlap of the observations and to disallow observations that are claiming the same proposal or detection.

The inference problem (\ref{equ:energy_observation_model}) is NP-hard.
We therefore obtain an approximate solution using the multi-branch method from \cite{Schindler06ECCV}.
Note that the obtained solution gives us a set of valid associations between the detections and 3D proposals.
As 3D proposals can mostly be obtained in the close camera range,
there will be detections that are not part of any selected observation.
For such detections, we augment our final observation set with \textit{partial observations}, containing detections, but no 3D information.
In particular, this helps to retain far-away targets which are not covered by 3D proposals.

The proposed approach of combining the information from different sources is also called \textit{early fusion}, as hard decisions are made prior to invoking the tracker.
This contrasts to \textit{late fusion}, where the tracker performs the selection and fusion of measurements.
However, \textit{late fusion} would result in a combinatorial explosion of the state-space of our multi-hypothesis tracker.
Additionally, the previously described selection of observations matches the selection process of tracking hypotheses through the tracker.
The scores and interaction terms consist of the same building blocks,
increasing the chance that the observations picked here will produce high-scoring hypotheses.

\section{Tracking}

Our tracking formulation follows a hypothesize-and-select framework, as initially proposed in \cite{Leibe08TPAMI},
which is a current state-of-the-art tracking paradigm for vision-based tracking \cite{Choi15ICCV}.
In this paradigm an over-complete set of hypotheses is created, and then the most suitable ones are selected.

In the following, we will first describe our 2D-3D Kalman filter.
It is applied to filter the observations which are associated to our trajectory hypotheses.
By generating an over-complete set of hypotheses we capture a multitude of possible data associations.

\subsection{Coupled Filtering of 2D-3D States}
We propose a formulation which keeps track of objects in both image domain and world-space.
In contrast, most state-of-the-art vision-based tracking methods perform tracking just in the image domain.

We use an Extended Kalman Filter (EKF) to estimate a joint 2D-3D state,
and couple the different quantities using projection and back-projection operations.
The geometry of a hypothesis is estimated using detections (\ref{equ:detection_obs}) and 3D proposals (\ref{equ:proposal_obs}),
which we filter using the EKF.
The state at time $t$ is defined by $\mathbf{x}_t = [ \hypobbox_2d, \hypobvelo_2d, \hypopose, \hypovelocity, \hyposize ]^T$.
For the position $\hypopose$ on the ground plane and the 2D bounding box $\hypobbox_2d$ we use a constant-velocity model,
which requires adding the rate of change of the bounding box $ \hypobvelo_2d $.

At each timestep, the bounding box position of each hypothesis is corrected for the ego-motion.
The footpoint of the bounding box $\hypobbox_2d$ is back-projected into world-space,
using the estimated distance from the camera computed using $\hypopose$.
The translation and rotation given by the ego-motion is then applied to the 3D point.
By projecting it back into the image, we obtain the corrected footpoint.

Then, the components of the position and bounding boxes are predicted using the corresponding velocities,
except for the following heights and the footpoint estimate, which are weakly coupled through these projection and back-projection operations:
\begin{eqnarray}
	\label{equ:couple:height3d}
	h^{3D} \! & \hspace{-7px} = \hspace{-7px} & w_{b} \tfrac{d_{c}}{f} h^{2D} \! + w_{a} h^{3D} \\
	\label{equ:couple:height2d}
	h^{2D} \! & \hspace{-7px} = \hspace{-7px} & w_{a} \left(\delta t \dot{h}^{2D} \! + h^{2D}\right) \! + w_{b} \tfrac{f}{d_{c}} h^{3d}.
\end{eqnarray}
\begin{eqnarray}
	\label{equ:couple:bb2d}
	y^{2D} \! & \hspace{-7px} = \hspace{-7px} & w_{a} \left(\delta t \dot{y}^{2D} \! + y^{2D}\right) \! + w_{b} \left(\tfrac{f}{d_{c}} \left(\delta{t} \dot{y}^{C} \! + y^{C}\right) \! + \! v_{0}\right) \\
	x^{2D} \! & \hspace{-7px} = \hspace{-7px} & w_{a} \left(\delta t \dot{x}^{2D} \! + x^{2D}\right) \! + w_{b} \left(\tfrac{f}{d_{c}} \left(\delta{t} \dot{x}^{C} \! + x^{C}\right) \! + \! u_{0}\right)
\end{eqnarray}
Here $\delta{t}$ is the time difference between two prediction steps, $f$ is the focal length, and $u_0, v_0$ are the horizontal and vertical components of the principal point.
The position in camera space $ \hypopose^{C} = \left[ x^C, y^C, z^C \right]^T $
helps to compute the distance from camera $d_c = z^C$.
The weights $w_{a}$ and $w_{b}$ determine how much the 2D and 3D states contribute to the coupling (we learn these weights on our training set, see Sec. \ref{sec:experimental}).

Depending on which observations are available (see \Sec~\ref{subsec:observation_fusion}), we perform sequential updates with these different sources of information.
We distinguish two cases:
(i) Fused observation, where a detection is associated to a 3D proposal,
resulting in observable values $\mathbf{z}_t = \left [ \hypobbox_2d, \hypopose, \hypovelocity, \hyposize \right ]^T$.
In this case we update the corresponding quantities of the state. The coupling only happens in the prediction step.
However, a measured bounding box will still influence the next 3D prediction through the coupling.
(ii) Partial observation (non-associated detections), which restricts the observable values to $\mathbf{z}_t = \left [ \hypobbox_2d, \hypopose_{bp}, \hyposize_{mean} \right ]^T$.
Here, we still update the 3D position using the back-projection of the footpoint of the bounding box $ \hypopose_{bp} $,
and we update the 3D size using the mean size $ \hyposize_{mean} $ of the category from the training data.
In this case, a different measurement variance is attached to the 3D position and size.

\subsection{Hypothesis Set Generation}
We perform tracking by maintaining an over-complete set of trajectory hypotheses $\hyposet = \left \{ \hypo{t_0:t_n}{k} \right \}$,
each defined on the time-span $t_0 \!\! : \!\! t_n$ (note that we will omit time indices where not needed).
Each hypothesis is built from the set of observations using our 2D-3D Kalman filter.
It is constrained to the ground plane and maintains a state estimate over time:
\begin{equation}
\label{equ:hypo_def}
\hypo{}{k} (t) = \left [ \hypobbox_2d, \hypobvelo_2d, \hypopose, \hypovelocity, \hyposize, \hypocategory, \obs{t}{k} \right ]^T.
\end{equation}
The observation used in each frame is also attached to the hypothesis,
although it may be missing for any given frame, \ie $ \obs{t}{k} = \emptyset $.
A hypothesis $\hypo{t_0:t_n}{k}$ is furthermore associated to its inlier set $\hypoinliers_k = \{ \obs{t}{k} | t_0 \le t \le t_n \}$ of observations spread out over the temporal domain.
We attach a multinomial distribution over possible object categories $\hypocategory = \left \{ \texttt{car}, \texttt{pedestrian}, \texttt{cyclist} \right \}$.
It is estimated from the associated detections by performing forward Bayesian filtering
(likelihood terms are learned from the data).


\PAR{Hypothesis Extension.}
We create an over-complete hypothesis set similar to the one proposed in \cite{Leibe08TPAMI}: at each time step,
we (i) extend existing hypotheses with a new observation
and (ii) generate alternative hypotheses starting at new observations within a temporal window.

In case several detections are very close to the hypothesis (in image-space, on the ground plane, and in appearance),
we branch the hypothesis, updating each branch with a different detection.



\PAR{Hypothesis Persistence.}
We stop updating hypotheses which have left the camera view frustum, but we keep extrapolating them for a while (see Fig.~\ref{fig:result_gct}).
This keeps the hypothesis `alive' and it continues to claim its inlier set during the optimization procedure,
such that these observations are not suddenly free to support other hypotheses which would be implausible otherwise.
In order to keep the size of the hypothesis set feasible, we remove duplicates and hypotheses which were not selected for some time.


\subsection{Hypothesis Selection}
In each frame we select from the set of hypotheses by performing MAP inference in a CRF model, similar to \cite{Leibe08TPAMI}:
\begin{equation}
\label{equ:energy_tracking_model}
\mathbf{E}(\binvec, \hyposet) = \sum_{h_i \in \hyposet} m_i \vartheta \left( h_i \right) + \sum_{h_i,h_j \in \hyposet} m_i m_j \psi \left( {h_i, h_j} \right ),
\end{equation}
where $\binvec \in \left \{ 0, 1 \right \}^{\left | \hyposet \right |}$ is a binary indicator vector,
with $m_i = 1$ meaning that a hypothesis has been picked.
We search for the selection $ \binvec^{*} = \argmin_{\binvec} \mathbf{E}(\binvec, \hyposet) $ with the lowest energy given a hypothesis set $ \hyposet $.

\PAR{Hypothesis Score.} The unary term scores a hypothesis:
\begin{eqnarray}
	\vartheta \left( \hypo{}{i} \right) = w^h_{min} -  \sum_{ \obs{t}{i} \in \hypoinliers_i} \hspace{-3px} \mathcal{S} ( \obs{t}{i}, \hypo{}{i}),
\end{eqnarray}
where the model parameter $w^h_{min}$ defines a minimal score for a hypothesis.
The contribution of an observation $\mathcal{S} \left ( \obs{t}{i}, \hypo{t_0:t_n}{i} \right)$ is: 
\begin{equation}
\mathcal{S} \left( \obs{t}{i} , \hypo{t_0:t_n}{i} \right ) = e^{\left( -\tau \cdot (t_n-t) \right)} \cdot s \left ( \obs{t}{i} \right ) \cdot \Phi \left ( \obs{t}{i} , \hypo{t_0:t_n}{i} \right ).
\end{equation}

Using an exponential decay, observations further in the past have less influence.
The score of an observation is $s \left ( \obs{t}{i} \right ) = s_{det}$. In the current formulation, we only use the score of the detection.
The association affinity term:
\begin{equation}
	\begin{aligned}
		& \Phi \left ( \obs{}{i} , \hypo{}{i} \right ) =  \mathbb{I}_c \left( \obs{}{i} , \hypo{}{i} \right ) \cdot \\
		& \left(
		w_c\Phi_{\text{c}} \left ( \obs{}{i} , \hypo{}{i} \right )
		+ w_m\Phi_{\text{m}} \left ( \obs{}{i} , \hypo{}{i} \right )
		+ w_p\Phi_{\text{p}} \left ( \obs{}{i} , \hypo{}{i} \right )
		\right)
		\end{aligned}
\end{equation}
is a linear combination of
the appearance score $\Phi_{\text{c}} \left ( \obs{}{i} , \hypo{}{i} \right )$,
the motion model term $\Phi_{\text{m}} \left ( \obs{}{i} , \hypo{}{i} \right )$,
and the projection model term $\Phi_{\text{p}} \left ( \obs{}{i} , \hypo{}{i} \right )$,
multiplied by an indicator function $\mathbb{I}_c \left(\cdot,\cdot \right )$ that prevents association between hypotheses and observations of incompatible categories.
The weights are functions of the distance $ d \left( \obs{}{i} \right) $ of an observation from the camera and the relative weighting of the appearance term $w_c$:
\begin{equation}
	w_m = ( 1 - w_c ) e^{\left (-\gamma \cdot d \left( \obs{}{i} \right ) \right) } \, , \: w_p = \left( 1-w_c-w_m \right ).
\end{equation}

For the appearance score we use intersection kernels over color histograms as in \cite{Choi15ICCV}.
The motion model term $\Phi_{\text{m}} \left ( \obs{}{i} , \hypo{}{i} \right ) \sim \mathcal{N} \left( \hypopose_{obs} | \hypopose^{h_k}_{pred}, \Sigma^{h_k}_{pred}\right )$
scores the probability of the observation given the Kalman filter prediction (on the ground plane).
The projection model $\Phi_{\text{p}} \left ( \obs{}{i} , \hypo{}{i} \right ) = \operatorname{IoU} \left( \hypobbox_2d \left( o\right ), \hypobbox_2d \left( h\right ) \right)$
is computed as the intersection-over-union between the predicted 2D bounding box of the hypothesis and the observed bounding box (in the image domain).

\PAR{Hypothesis Interaction.} The pairwise potential of the CRF in (\ref{equ:energy_tracking_model}) scores the interaction of each pair of hypotheses:
\begin{eqnarray}
	\psi \left( {h_i, h_j} \right ) = \, w^h_{ol} \,  O(\hypo{}{i}, \hypo{}{j}) +
		w^h_{sh} \left| \hypoinliers_i \cap \hypoinliers_j \right|.
\end{eqnarray}
The parameter $w^h_{ol}$ weights the physical overlap penalty:
\begin{equation}
	O \left( \hypo{}{i}, \hypo{}{j} \right) = \sum_t \left[ \operatorname{IoU} \left( \hypobbox_2d \left( \hypo{}{i}, t \right), \hypobbox_2d \left(  \hypo{}{j}, t \right) \right) \right]^2,
\end{equation}
which punishes overlap in image space of the two hypotheses.
Additionally, we add a penalty $w^h_{sh}$ for each observation shared by the hypotheses.

Intuitively, after solving this inference problem we obtain a set of best-scoring trajectories that are physically plausible (\ie do not overlap in space-time).
Note the similarity of this model to the observation fusion in Sec.~\ref{subsec:observation_fusion}.
The difference is that in this case we are aiming for a partition of the observations over time,
whereas in Sec.~\ref{subsec:observation_fusion} we are computing associations between 3D proposals and detections.

\section{Experimental Evaluation}
\label{sec:experimental}

All our evaluations are based on the KITTI Vision Benchmark Suite \cite{Geiger12CVPR}.
It provides training sequences with a publicly available ground truth, and a separate test set which can only be evaluated using the provided
evaluation server.
Evaluations are based on the CLEAR MOT metrics \cite{Bernardin08JIVP}.
KITTI furthermore provides detections for cars, pedestrians, and bicycles from two different detectors \cite{Wang13ICCV, Geiger2014PAMI}.
In our experiments we use the detections from the Regionlet detector \cite{Wang13ICCV}.
We split the KITTI training data into two disjoint sets: a training set which is used for optimizing parameters (based on MOTA score) via Hyperopt \cite{Bergstra13ICML} and a validation set for evaluating different aspects of our pipeline\footnote{
We used sequences 1,2,5,7,8,9,11,17,18,19 as training set and sequences 0,3,4,6,10,12,13,14,15,16,20 to perform validation.}.
We interpret the reported result mostly based on the achieved MOTA score, which is considered to be the most distinctive tracking evaluation measure~\cite{breuersWACV16}.



%

%
%


\PAR{Observation Precision.}
For our detailed analysis of the fused observations, we use 2D and 3D annotations provided in the KITTI training set.
We focus on single frame results here, in order to gain insights into the observation fusion without regarding the tracking.

\begin{figure}[t]
\begin{center}
\includegraphics[width=0.49\linewidth]{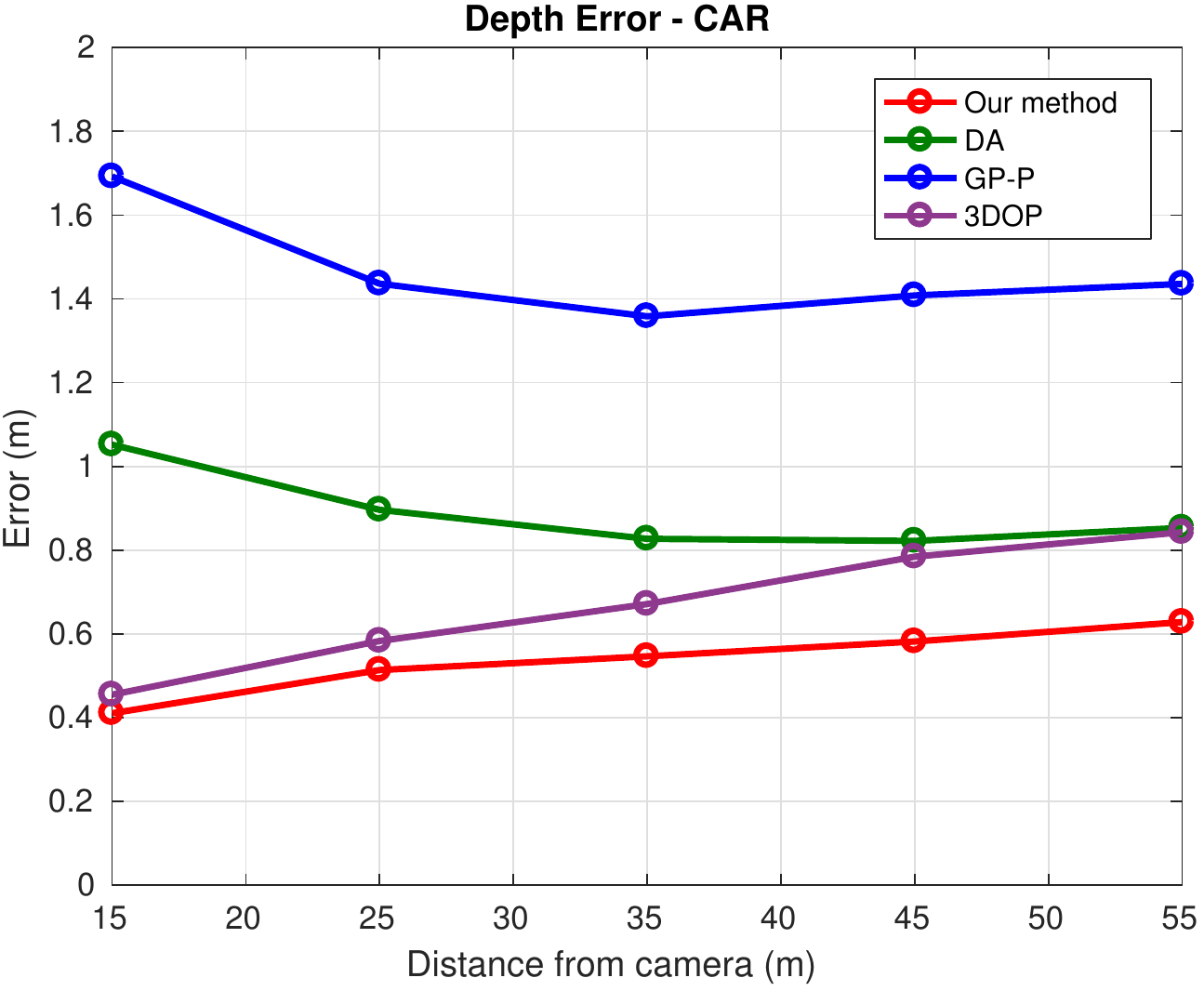}
\includegraphics[width=0.49\linewidth]{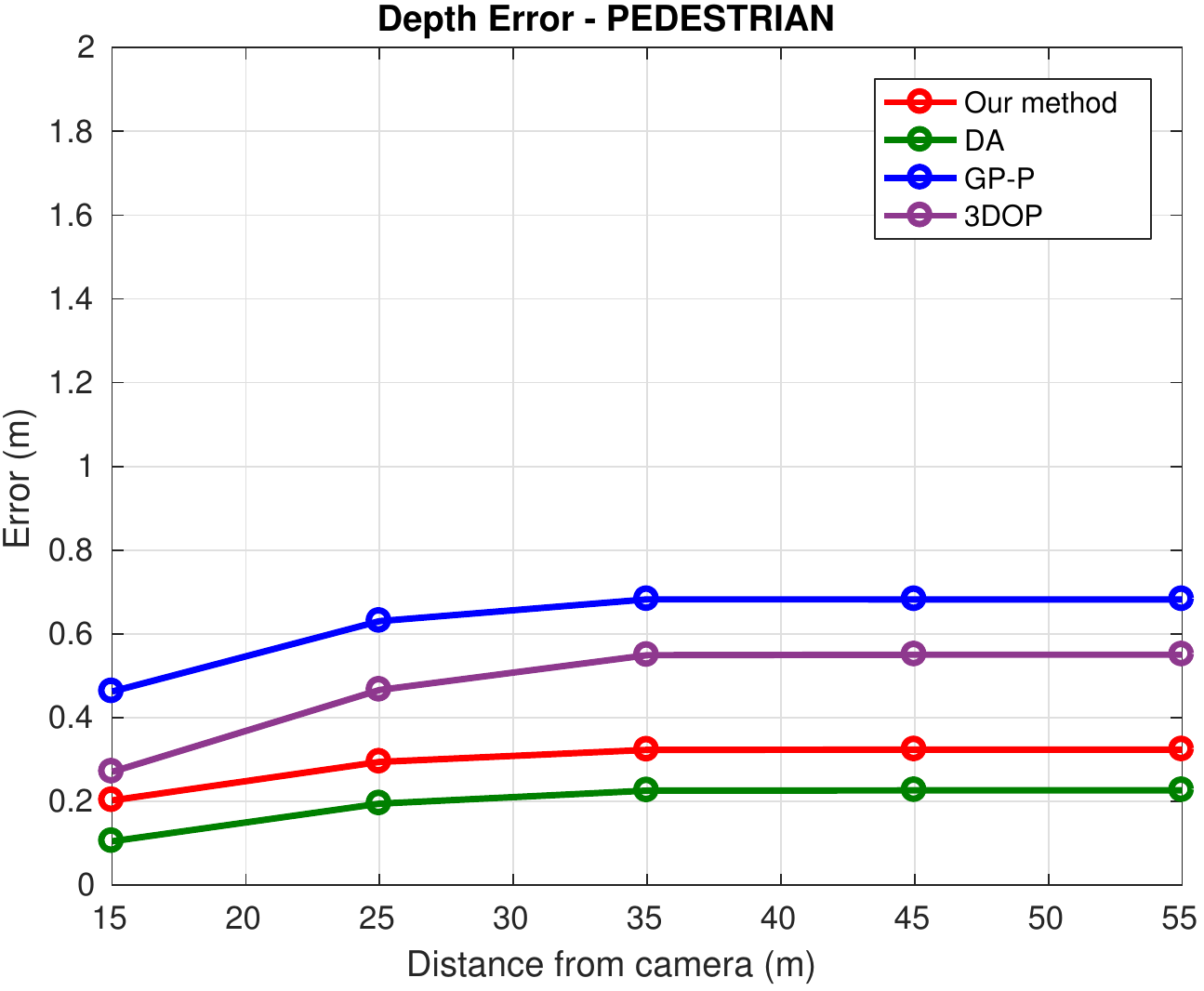} \\[7pt]
\includegraphics[width=0.49\linewidth]{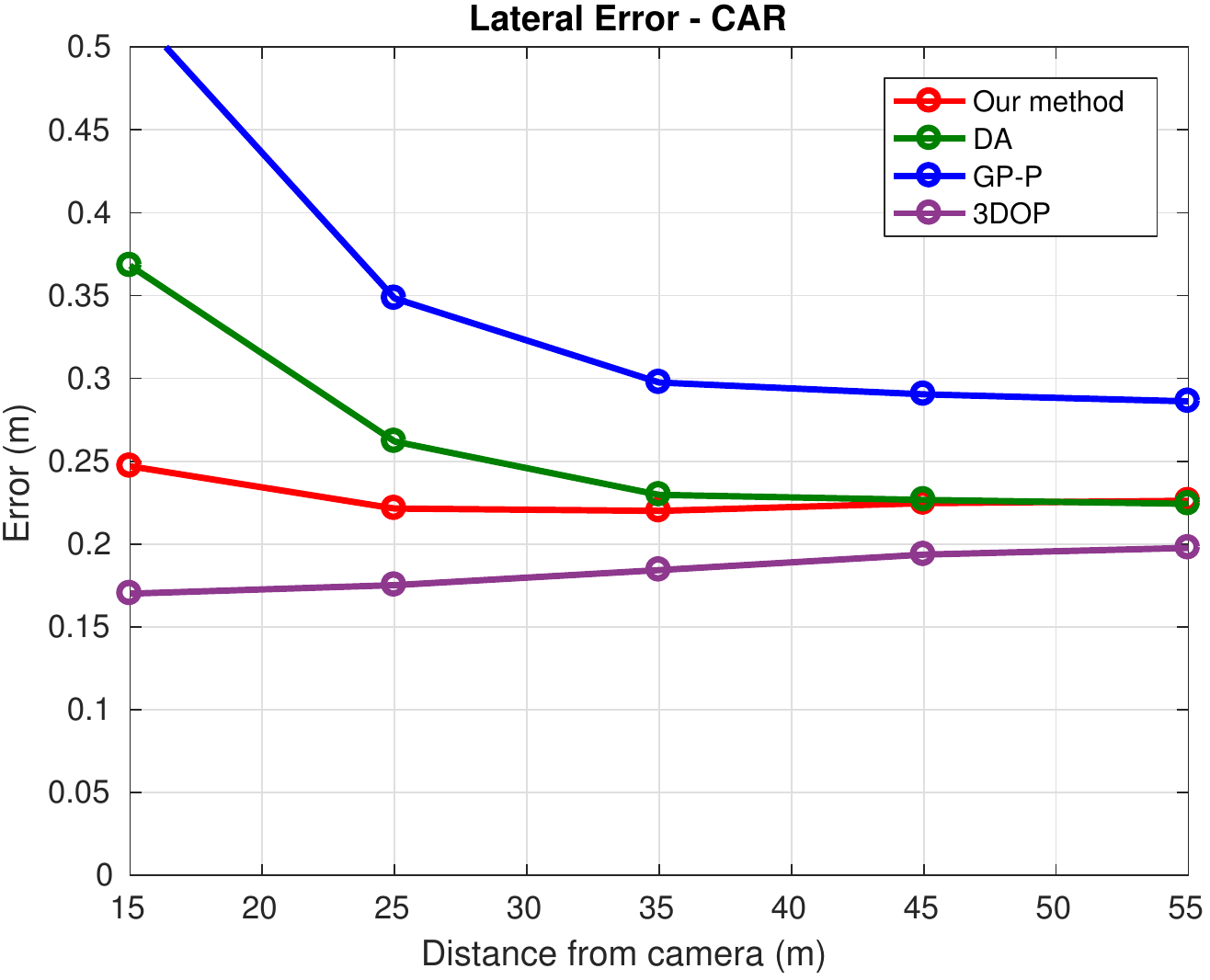}
\includegraphics[width=0.49\linewidth]{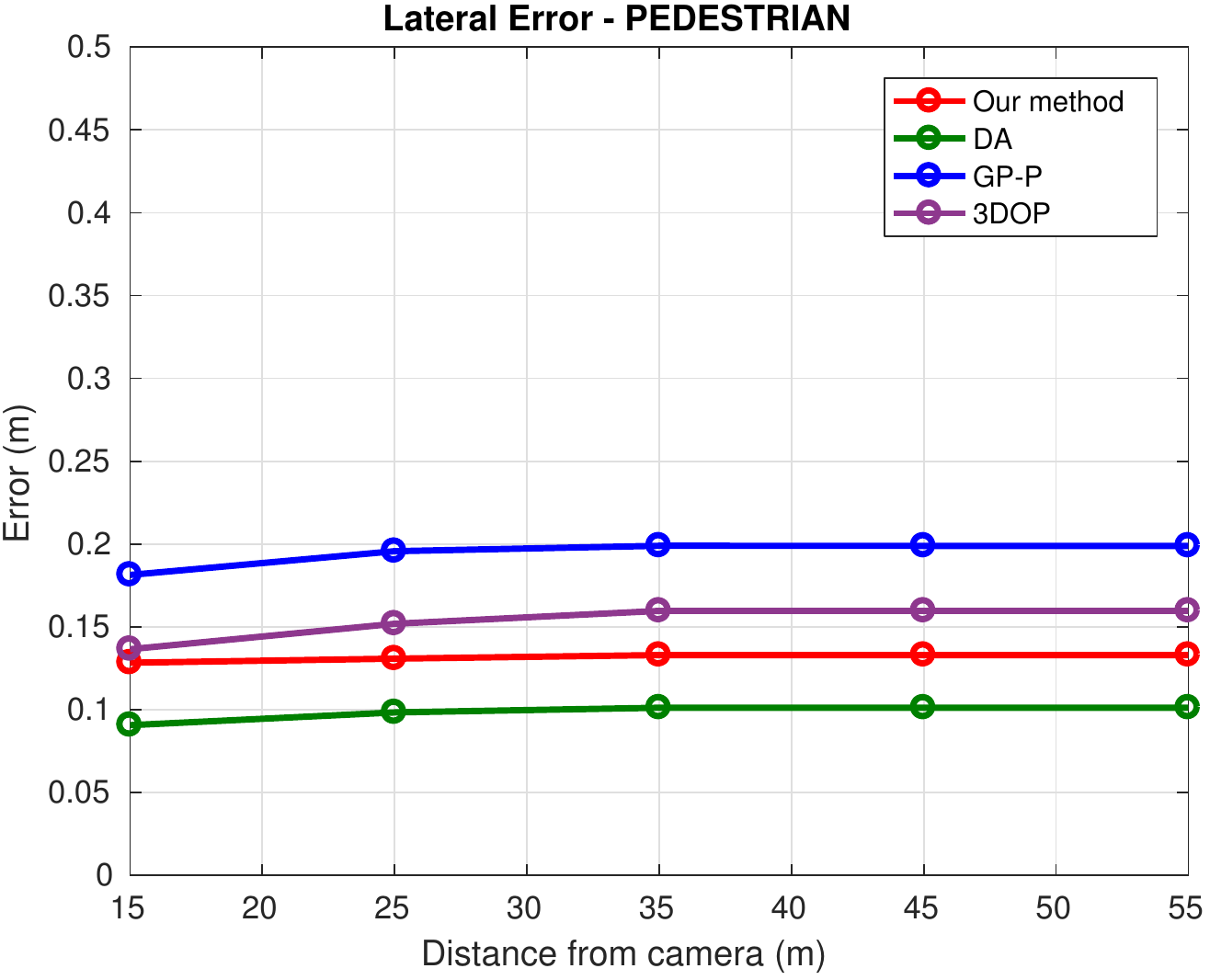}
\end{center}
\vspace*{-12pt}
\caption{Localization error by distance range in depth and lateral direction.}
\label{fig:position_error_by_range}
\end{figure}

\Fig~\ref{fig:position_error_by_range} shows the positioning error by distance range,
split into the error made in estimating the distance to the camera,
and the lateral error orthogonal to the depth error.
We compare the performance of our 3D proposals to three baselines:
(GP-P) Ray-casting of the footpoints of the 2D bounding boxes and intersection with the ground plane;
(DA) Results obtained by depth analysis \cite{Ess09PAMI};
(3DOP) Results from the recent 3DOP \cite{Chen15NIPS}.
While 3DOP has ultimately other goals, the estimation of 3D bounding boxes is an essential step in the pipeline.

For cars, our evaluation shows that simpler methods (GP-P and DA) are clearly outperformed by the more sophisticated ones. 
The lateral position of cars is best estimated by 3DOP, which is specially designed for estimating bounding boxes of specific object categories, while our method is more general.
Our method is one of the two best performing methods in all cases. While DA is performing well on the \textit{pedestrian} category, it lacks precision in the \textit{car} category;
the situation is reversed for 3DOP.

\PAR{Ablation Study.}
In order to evaluate the influence of the different ingredients of our approach, 
we switch off parts of our pipeline.
MOTA and MOTP are both computed using the 2D bounding box overlap, which is the standard in KITTI.
See Table~\ref{fig:tracking_results_baseline} for the results:
(No-flow) uses no scene flow;
(Det. only) does not use 3D proposals;
(2D-tracker) is a pure 2D version of our tracking pipeline, disabling even the ground plane estimation and visual odometry;
(DA) uses depth analysis \cite{Ess09PAMI} rather than our 3D proposals to obtain the 3D measurements.
As can be observed, in this 2D evaluation our full method performs best, and each of the components contributes to the performance.
A clear benefit of our method will be seen when evaluating in world space.

\begin{table*}[t]
\begin{center}

\caption{Ablation study.}
\label{fig:tracking_results_baseline}
\newlength{\thiscol}
\setlength{\thiscol}{6mm}
\newlength{\thiscolS}
\setlength{\thiscolS}{4mm}

\setlength{\thiscol}{6mm}
\setlength{\thiscolS}{4mm}

\begin{tabular}[]{|p{15mm}|p{\thiscol}p{\thiscol}p{\thiscolS}p{\thiscolS}p{\thiscol}p{\thiscol}p{\thiscol}|}
\hline
 \textbf{Cars}   & MOTA  & MOTP  & ID  & Frag & MT    & PT    & ML \\
\hline
 Full Version    & \textbf{74.38} & \textbf{82.85} & 26  & 131  & \textbf{49.59} & 40.68 & 9.80 \\
 No-flow         & 74.17 & 82.74 & 31  & 141  & 49.50 & 40.20 & 10.29 \\
 Det\onedot only & 73.99 & 82.66 & 48  & 152  & 49.01 & 40.19 & 10.78 \\
 2D-tracker      & 72.29 & 82.40 & \textbf{11}  & \textbf{72}  & 43.13 & 42.64 & 14.22 \\
\hline
 DA              & 72.93 & 82.56 & 108  & 201  & 49.50 & 37.25 & 13.24 \\
\hline
\end{tabular}
\begin{tabular}[]{|p{15mm}p{\thiscol}p{\thiscol}p{\thiscolS}p{\thiscolS}p{\thiscol}p{\thiscol}p{\thiscol}|}
\hline
 \textbf{Pedestrians} & MOTA  & MOTP  & ID  & Frag & MT    & PT    & ML \\
\hline
 Full Version         & \textbf{61.87} & 78.85 & 41  & 164  & \textbf{55.95} & 33.33 & 10.71  \\
 No-flow              & 61.82 & 78.89 & 53  & 175  & 54.76 & 34.52 & 10.71 \\
 Det\onedot only      & 61.13 & 78.88 & 51  & 172  & 55.95 & 34.52 & 9.52  \\
 2D-tracker           & 59.74 & 78.85 & 59 & 162  & 48.81 & 35.71 & 15.47 \\
\hline
 DA                   & 61.69 & \textbf{78.97} & \textbf{32} & \textbf{148}  & 55.95 & 33.33 & 10.71 \\
\hline
\end{tabular}

\end{center}
\end{table*}


\PAR{3D Localization Evaluation.}
%
For each true positive association we evaluate the distance in 2D (as intersection-over-union) and in 3D (as the Euclidean distance on the ground plane).
This allows to compute the MOTP-2D and MOTP-3D metrics as suggested by \cite{Bernardin08JIVP}:
\begin{eqnarray}
	\text{MOTP-2D} & = & \frac{ \sum_{i,t} \operatorname{IoU} \left( \mathbf{b}^i_{gt}(t), \mathbf{b}^i_{traj}(t) \right) }{ \sum_t n_{tp}(t) } \\
	\text{MOTP-3D} & = & \frac{ \sum_{i,t} \left\| \hypopose^i_{gt}(t) - \hypopose^i_{traj}(t) \right\| }{ \sum_t n_{tp}(t) },
\end{eqnarray}
with the number of true positive associations per frame $ n_{tp} $.

\begin{figure}[t]
\begin{center}
\includegraphics[width=0.49\linewidth]{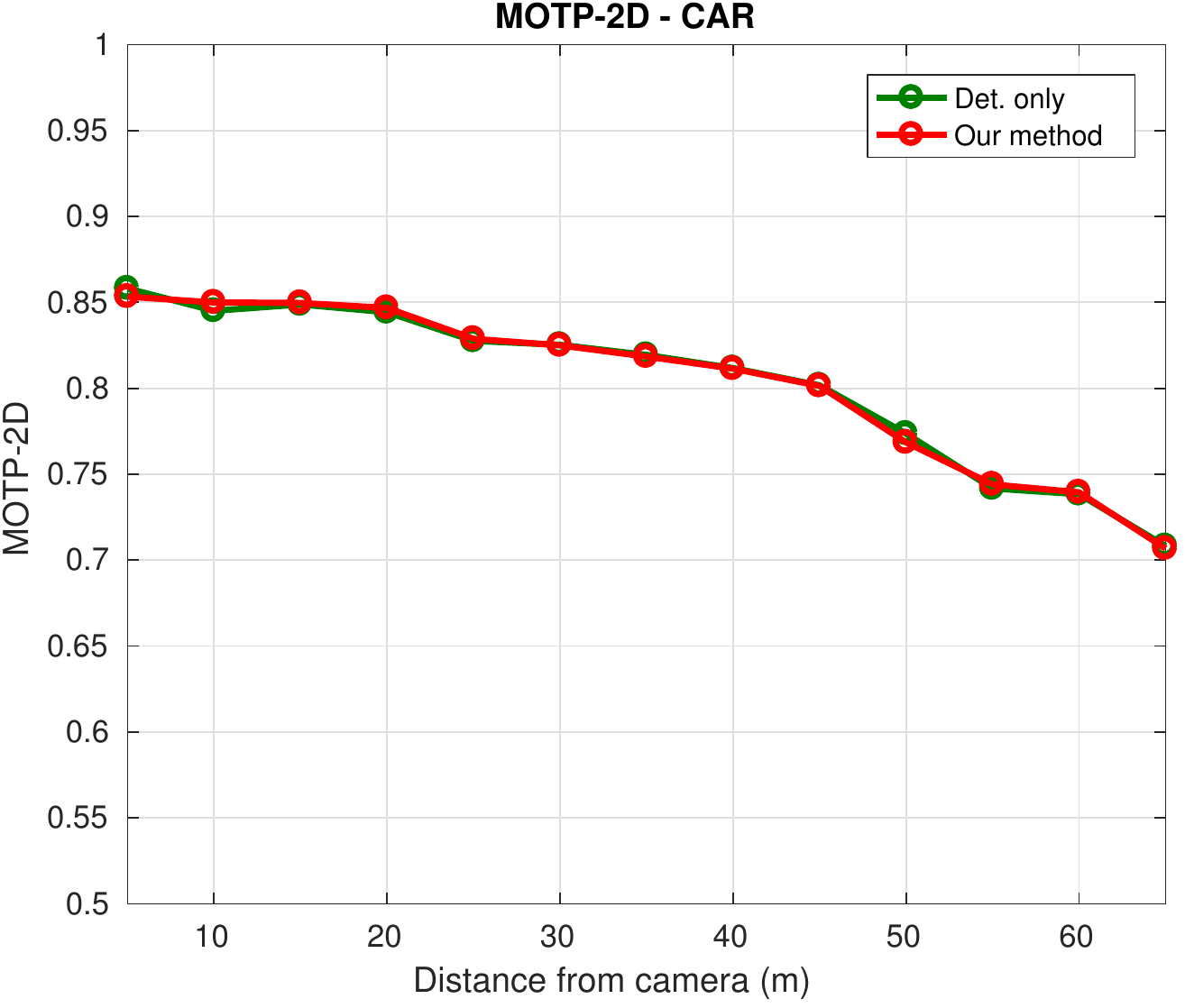}
\includegraphics[width=0.49\linewidth]{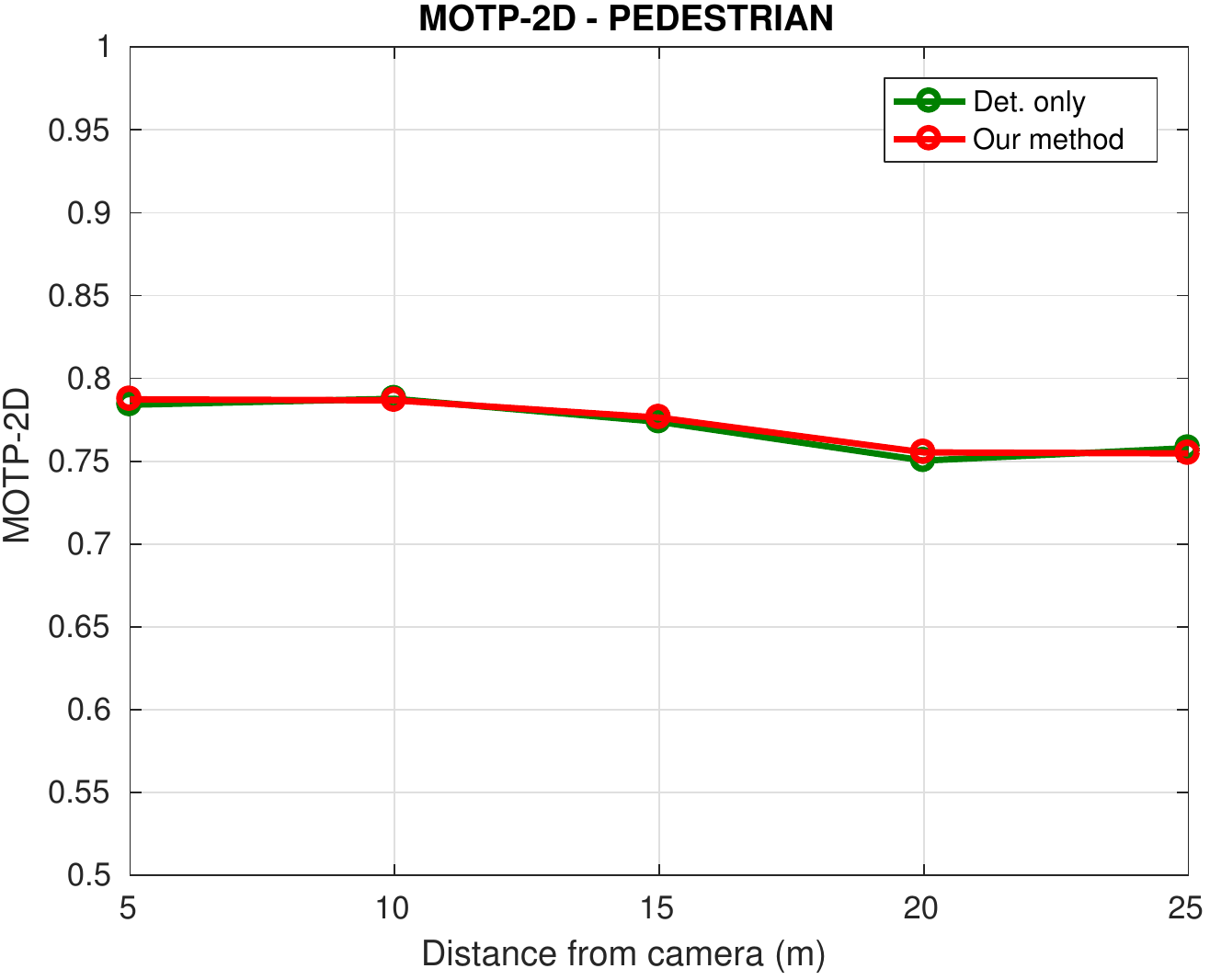} \\[7pt]
\includegraphics[width=0.49\linewidth]{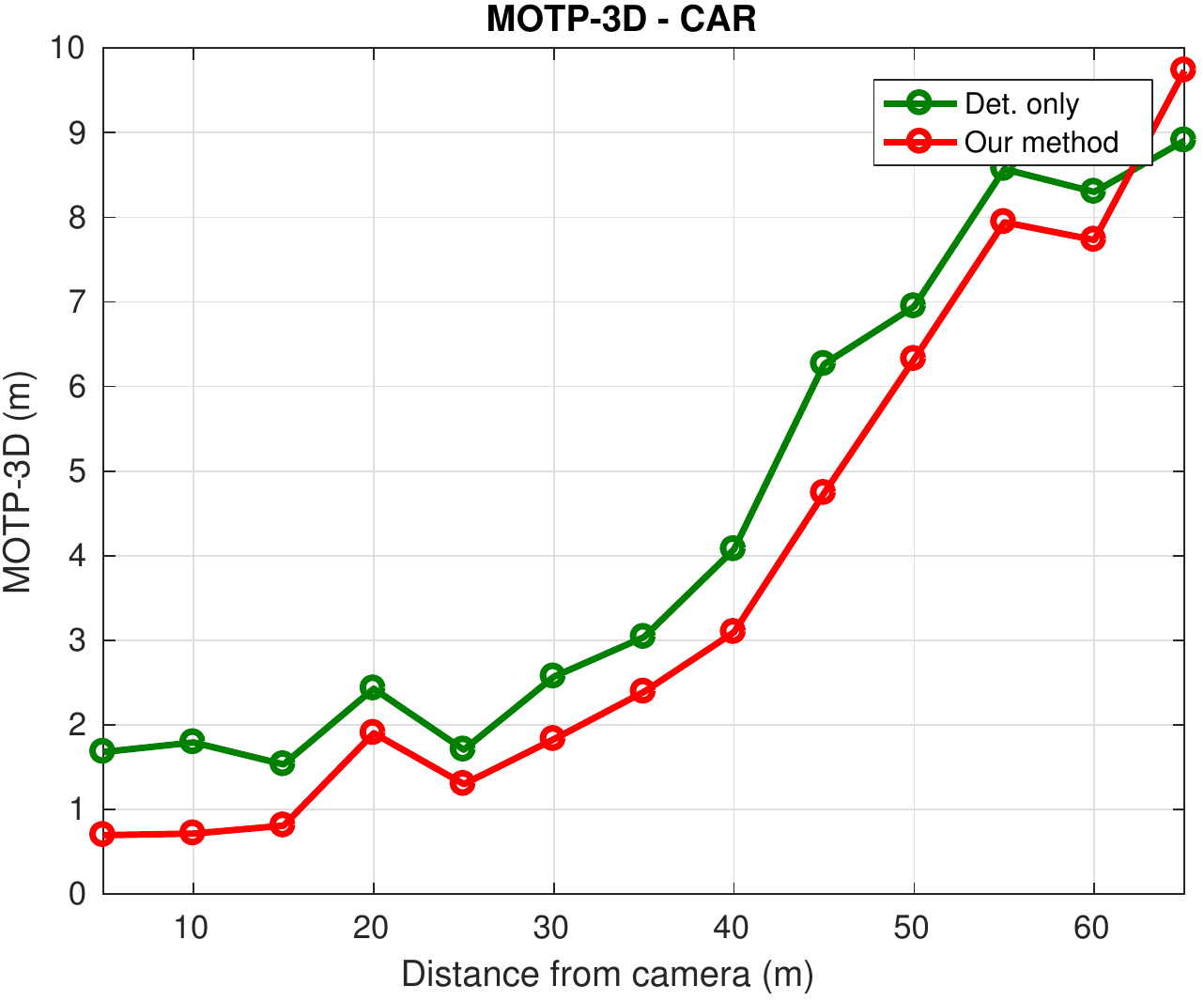}
\includegraphics[width=0.49\linewidth]{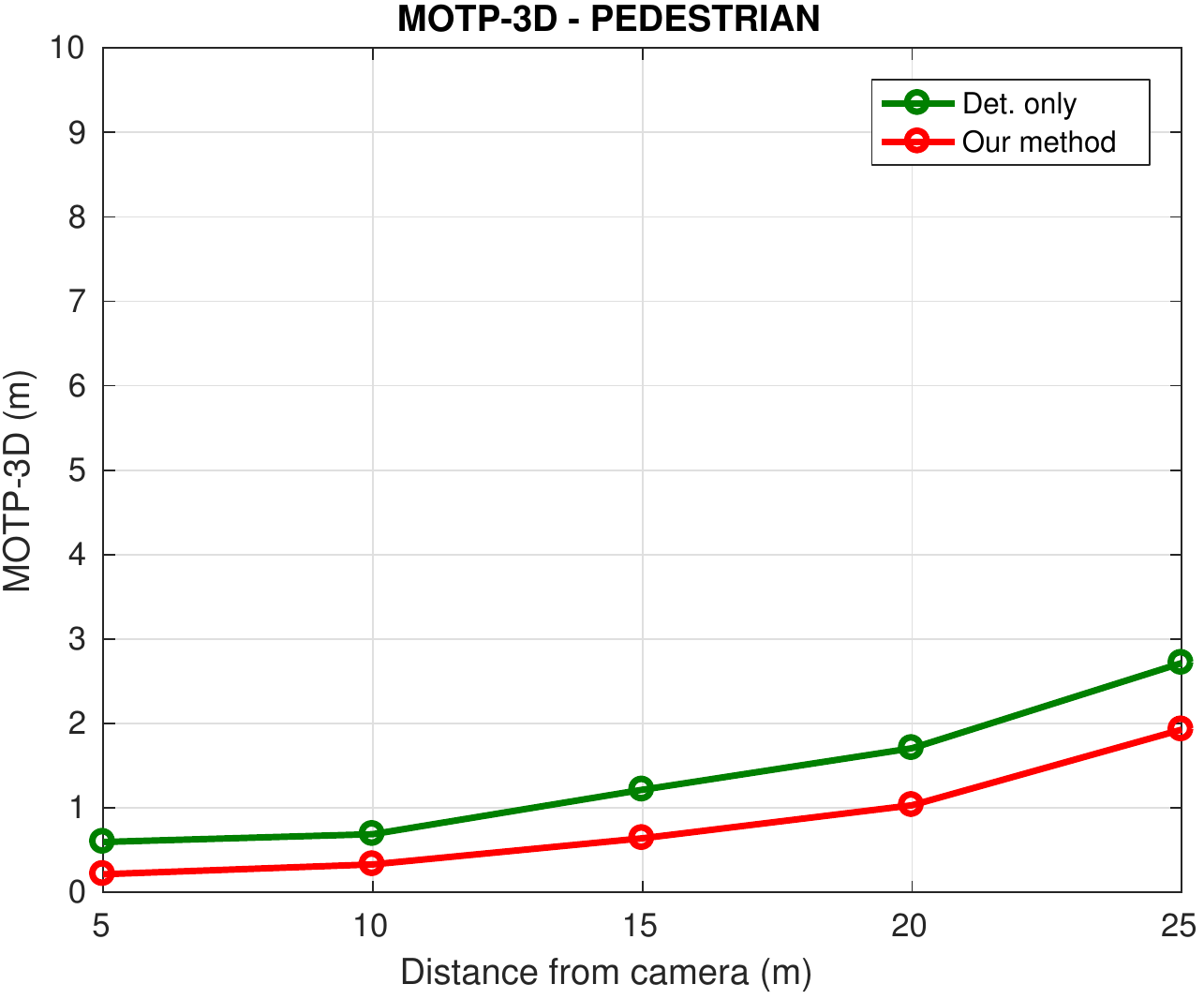}
\end{center}
\vspace*{-12pt}
\caption{Tracking Precision MOTP by distance range in 2D (higher values are better) and 3D (lower values are better) for cars and pedestrians.}
\label{fig:motp_by_range_comparison}
\end{figure}

We compare the results using both metrics in \Fig~\ref{fig:motp_by_range_comparison}.
For MOTP-2D (higher values are better), there is barely any difference between the full version and the detection-only baseline.
However, for MOTP-3D (lower values are better), 3D localization precision is considerably better.
This experiment shows that one can benefit from using 3D information in vision-based tracking,
even without compromising the ability to accurately track objects in the image domain.
Given the considered applications, the MOTP-3D results are clearly more relevant,
since in practical tasks we want to obtain precise information in the real world, not in image space.

\begin{figure}[t]
\includegraphics[width=0.49\linewidth]{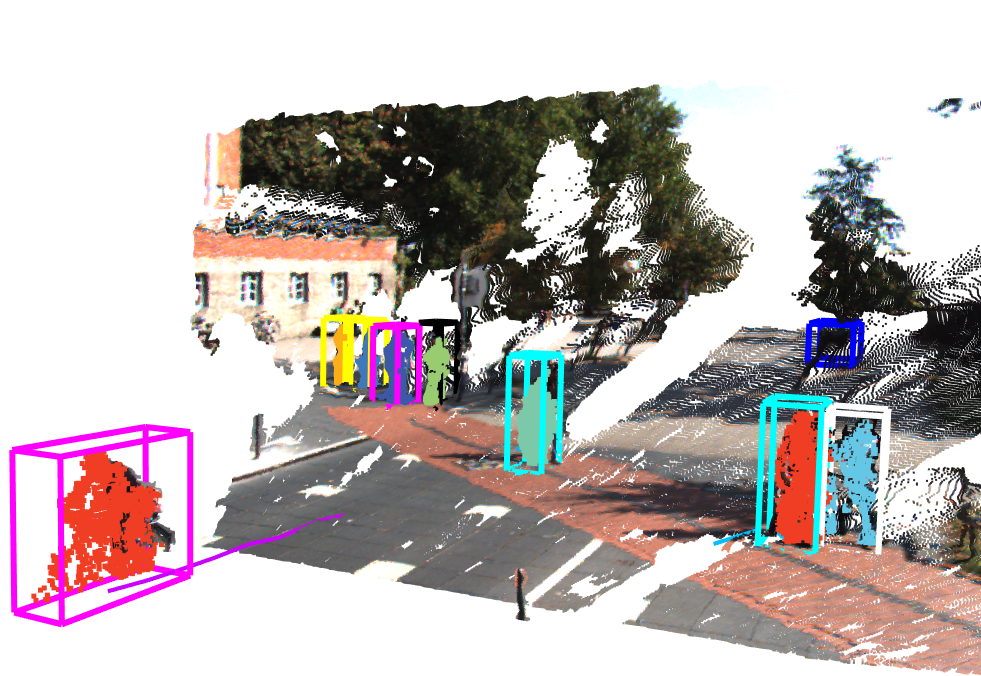}
\includegraphics[width=0.49\linewidth]{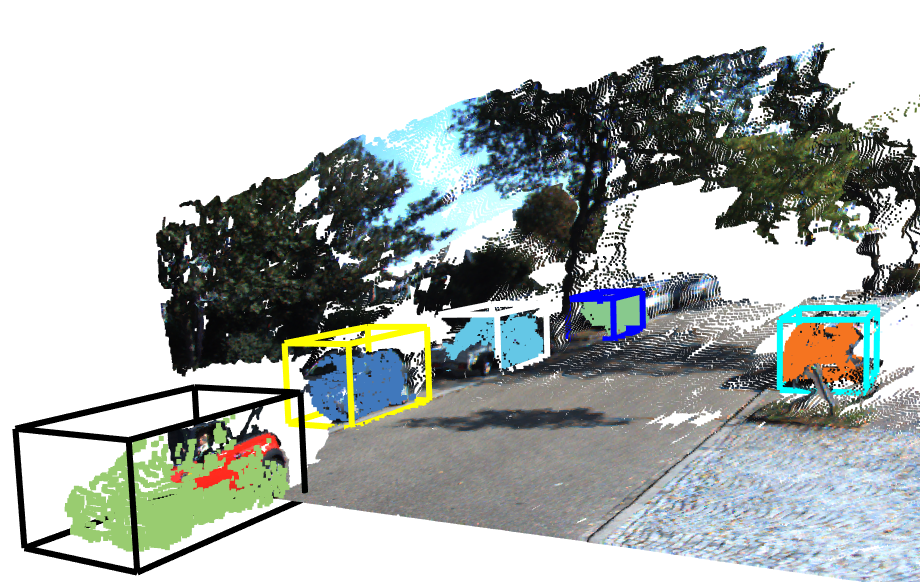}
\caption{Precise 3D localization enables us to do shape integration, and increase the system's awareness of its surroundings. \textit{Best viewed in color.}}
\label{fig:result_gct}
\end{figure}

\PAR{Exploiting Precise 3D Segmentations.}
As has been shown, our approach is suitable for performing precise 3D localization of tracked objects.
This localization can be utilized to accumulate 3D measurements over time, in order to reconstruct more of the shape of tracked objects than can be seen in one frame.
The proposals come with precise segmentations of the 3D point cloud (compare Fig.~\ref{fig:bb_and_3d_proposals}).
The precision of these segmentation masks results in a clean shape representation.
The duration of tracking helps to accumulate more depth data over time.
In Fig.~\ref{fig:result_gct} we accumulate measurements by using the GCT representation and weighted ICP proposed by~\cite{Mitzel12ECCV}.
Alternatively, one could use the segments as an input to~\cite{Held14RSS}.
The provided result can only be obtained when performing precise segmentation in 3D.
As objects and possible occluders can be well separated in world space, the shape of the tracked objects can be acquired with less noise.

\PAR{KITTI Evaluation Results.}
Table~\ref{fig:tracking_results_kitti} shows the result of evaluating our full pipeline on the official KITTI test set \wrt the highest-ranked baselines (note, that for NOMT \cite{Choi15ICCV}, we used the online version, NOMT-HM). Although we perform tracking jointly in 2D and 3D, the official KITTI evaluation is based on bounding box overlap in the 2D image domain. 

\begin{figure}[t!]

\includegraphics[width=0.98\linewidth]{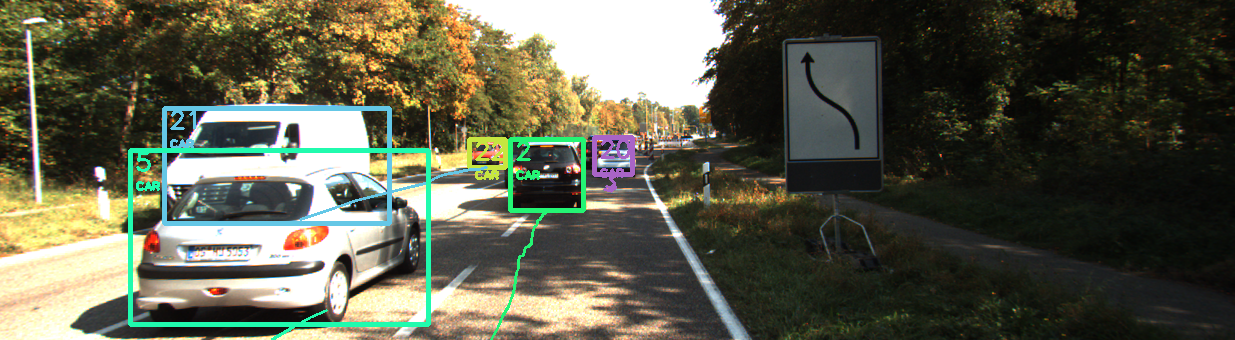}
\includegraphics[width=0.98\linewidth]{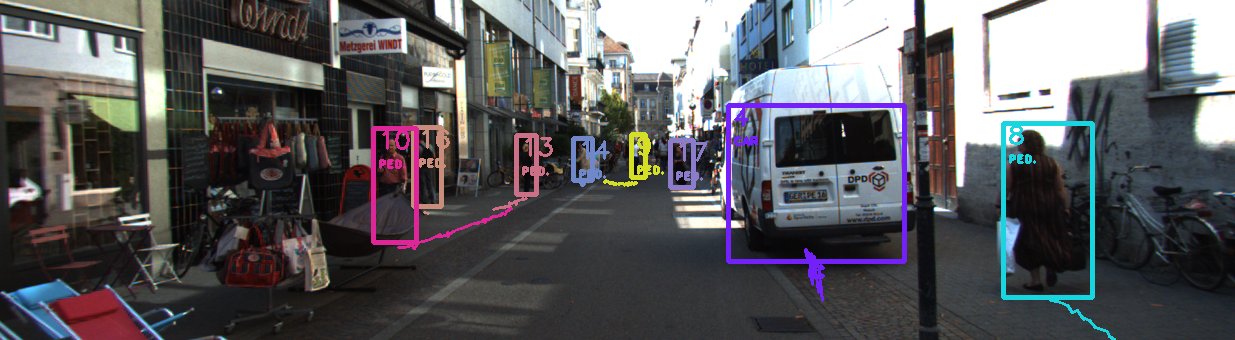}
\includegraphics[width=0.98\linewidth]{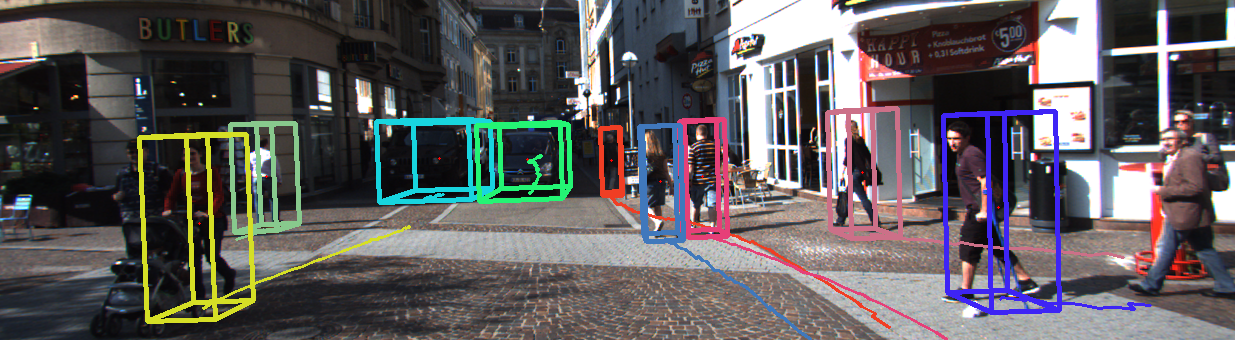}
\includegraphics[width=0.98\linewidth]{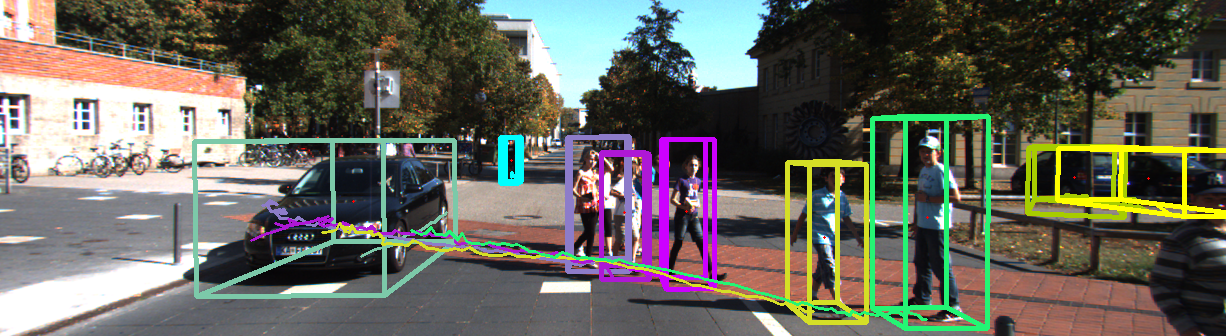}

\caption{Qualitative results on the KITTI tracking dataset showing 2D and 3D bounding boxes of the tracked objects.}
\label{fig:tracking_qualitative}
\end{figure}

We achieve highly competitive results in both categories, cars and pedestrians. For cars, our result is on par with the best performer, NOMT-HM~\cite{Choi15ICCV}. In contrast, for pedestrians we clearly outperform NOMT-HM. Similar to us, SCEA~\cite{Yoon16CVPR} has consistently good results on both categories and is the top performer for the pedestrians category, with our approach a close second.
Fig.~\ref{fig:tracking_qualitative} shows a selection of qualitative tracking results.

\begin{table*}[t]
\begin{center}

\caption{Results on the Kitti Benchmark. Tracking Accuracy (MOTA) and Precision (MOTP),
	ID-switches (ID), Fragmentations (Frag)
	Mostly Tracked (MT), Partly Tracked (PT), Mostly Lost (ML).}
\label{fig:tracking_results_kitti}

\setlength{\thiscol}{6mm}
\setlength{\thiscolS}{4mm}

\begin{tabular}[]{|p{18mm}|p{\thiscol}p{\thiscol}p{\thiscolS}p{\thiscolS}p{\thiscol}p{\thiscol}p{\thiscol}|}
\hline
 \textbf{Cars} & MOTA & MOTP & ID & Frag & MT & PT & ML \\
\hline
	Our method                & \textit{67.35} & 79.25          & 169          & 675          & 48.93          & 40.09 & 10.98 \\
	NOMT-HM \cite{Choi15ICCV} & \textbf{67.92} & \textbf{80.02} & 109          & \textbf{371} & 49.24          & 37.65 & 13.11  \\
	SCEA \cite{Yoon16CVPR}    & 67.11          & \textit{79.39} & \textit{106} & \textit{466} & \textit{52.13} & 36.89 & 10.98 \\
	LPSSVM \cite{Wang15BMVC}  & 66.35          & 77.80          & \textbf{63}  & 558          & \textbf{55.95} & 35.82 & 8.23 \\
	mbodSSP \cite{Lenz15ICCV} & 62.64          & 78.75          & 116          & 884          & 48.02          & 43.29 & 8.69 \\
	RMOT \cite{Yoon15WACV}    & 53.03          & 75.42          & 215          & 742          & 39.48          & 50.46 & 10.06 \\
\hline
\end{tabular}
\begin{tabular}[]{|p{15mm}p{\thiscol}p{\thiscol}p{\thiscolS}p{\thiscolS}p{\thiscol}p{\thiscol}p{\thiscol}|}
\hline
 \textbf{Pedestrians} & MOTA & MOTP & ID & Frag & MT & PT & ML \\
\hline
  Our method & \textit{38.37} & \textit{71.44} & 113         & 912          & 13.40          & 51.55 & 35.05 \\
  NOMT-HM    & 31.43          & 71.14          & 186         & 870          & \textbf{21.31} & 36.77 & 41.92 \\
  SCEA       & \textbf{39.34} & \textbf{71.86} & \textbf{56} & \textbf{649} & 16.15          & 40.55 & 43.30 \\
  LPSSVM     & 34.97          & 70.48          & \textit{73} & 814          & \textit{20.27} & 45.36 & 34.36 \\
  mbodSSP    & -              & -              & -           & -            & -              & -     & - \\
  RMOT       & 36.42          & 71.02          & 156         & \textit{760} & 19.59          & 39.18 & 41.24 \\
\hline
\end{tabular}

\end{center}
\end{table*}

\PAR{Runtime.}
Our full tracking pipeline requires 347~ms per frame, excluding external components (Intel I7 CPU, single thread, not optimized).
When not using observation fusion and the corresponding 3D proposals, each frame takes 48~ms.

\section{Conclusion}
We presented a novel tracking pipeline which combines 2D and 3D measurements. Our approach shows promising results, follows a clean design and is easily extendable.
In future work, we plan to exploit class-agnostic tracking of objects originating from our 3D object proposals.
In the context of autonomous driving cars and pedestrians are the most prominent categories, but other obstacles, from potentially unknown categories, should be identified as well.


\PAR{Acknowledgements}
This work was funded by ERC
Starting Grant project CV-SUPER (ERC-2012-StG-307432).

{\small
\bibliographystyle{ieee}
\bibliography{abbrev_short,egbib}
}

\end{document}